# Application of Algorithms in Energy-Efficient Design Platforms for Green Building

Na Yu[a], Fu Wenli[a] and Guo Fei[a]
[a]First Highway Engineering Co., Ltd. Chengdu, Sichuan Province, 610000, China

**Abstract.** During the green building design phase, computer-aided energy performance assessment has been adopted to achieve efficiency improvement and overall optimization. This paper constructs a platform based on powerful algorithms, combining Building Information Modeling (BIM) with sensor-generated operational data and high-end simulation optimization workflows. The multi-layer service architecture of the platform consists of dynamic energy simulation modules and evolutionary multi-objective optimization, which are closely connected through a high-performance C++ core and adaptive agent models. The mid-rise office building serves as the case study subject. In order to collect information on the characteristics of the building envelope and occupancy, five representative areas were selected. After preprocessing, the missing sensor data accounted for 3.2% of the total annual data, and all analysis variables were standardized through 15-minute interpolation. After 40 rounds of optimization, the annual energy consumption per square meter of the platform increased by 29.3% compared to the initial value, decreasing from 315 kWh/m² to 223 kWh/m². The increase in the lifecycle cost for residents is limited to within 3.7%, and the number of discomfort hours is reduced to less than 70 hours per year. According to the analysis of the Pareto optimal solution, the U-value of the envelope structure is 1.05-1.57 W/m²K, and the nighttime ventilation rate is 2.1-3.6 $h^{-1}$, which is closely related to energy performance. According to the above results, the algorithm integration for green building design indicates scalability and performance, as well as technical and engineering feasibility. Provide a reliable decision support tool for design engineers and sustainability practitioners, helping them build energy-efficient optimized buildings in an accurate and data-driven manner.

## 1. Introduction

Due to global urbanization and rapid economic growth, the energy consumed by the construction industry currently accounts for more than one-third of the total global energy demand [1]. With the increasing risks of climate change and the scarcity of resources, Chinese society is placing greater emphasis on improving the energy efficiency of buildings [2]. Most green building practices believe that using energy-saving technologies and new-generation buildings can reduce building emissions and their impact on the environment [3]. Due to the interconnection between building systems, occupant behavior, climate, and other regulations, achieving energy savings is challenging [4].

To address these issues, many researchers are studying the introduction of advanced algorithms into the design platforms of energy-saving devices [5]. Simulation analysis, optimization, and computational intelligence can be used to explore large-scale design spaces, simultaneously addressing multiple performance objectives and providing quantifiable support for the long-term management of buildings at different stages [6]. Advanced design algorithms such as multi-objective optimization, machine learning-assisted modeling, and parametric simulation can be used to improve the accuracy and efficiency of building energy consumption assessments, and provide customized energy-saving strategies for complex urban areas [7]. Algorithm-based platforms can also help automate tasks that were previously done manually, promote interdisciplinary collaboration, and disseminate best practices in green building engineering [8].

This paper will study the application of algorithms in energy-efficient building design platforms. In order to deeply study how advanced computational methods can accelerate the development and application of green building technologies, this paper provides a detailed analysis of platform structure, algorithm selection and integration, and their impact on actual design projects. These findings provide researchers and professionals with new methods and applications to promote the integration of intelligent design tools with green building.

## 2. Related Work

In the past decade, the development of green building energy-saving design platforms has been relatively rapid due to the increasing demand for environmental protection and the push from new regulations. Early digital platforms were primarily used to add basic energy simulation features to existing design tools; however, they lacked automation and analytical support for complex design choices [9]. As the demand for energy-efficient and long-term operational buildings increases, platform structures are becoming increasingly advanced. Many platforms now use BIM, parametric design, and semantic data integration [10]. Due to open frameworks and interoperable standards, the simulation engines, material databases, and regional compliance requirements supported by these platforms have increased. This has increased the variety of design scenarios and ensured robust regulatory compliance [11]. In summary, the aforementioned improvements lay a relatively strict, flexible, and collaborative foundation for the practice of green building design [12].

The foundation of these platforms is algorithms, which achieve energy and sustainability goals through intelligent control, simulation, and optimization. In the past, deterministic and stochastic optimization techniques (such as particle swarm optimization, genetic algorithms, and simulated annealing) were the primary methods for studying the complex nonlinear and high-dimensional design spaces of modern buildings [13]. On the other hand, many high-fidelity simulation tools, such as EnergyPlus and TRNSYS, are now often used together to conduct in-depth performance evaluations under various conditions, such as climate fluctuations and changes in occupancy and usage patterns. In recent years, the speed of performance prediction, anomaly detection, and adaptive design recommendations based on historical and real-time data has been supported by machine learning and artificial intelligence technologies [14]. On the other hand, data-driven and simulation-based methods are gradually merging to expand the field of green building technologies and applications [15].

However, so far, the developed research and platforms have been less than satisfactory. Computational optimization usually requires more resources and may not converge in large, multi-objective projects. Simulation-centered workflows are relatively slow and not suitable for early feasibility analysis and iterative prototype design. In the absence of or biased data, even AI-based strategies, whether automated or adaptive, can lead to inaccuracies or overfitting of training data. In the future, the industry will adopt more hybrid algorithm frameworks to promote richer data flow and sharing, and encourage interdisciplinary collaboration to design green buildings.

## 3. Methodology

### *3.1. System Architecture and Workflow*

The new energy-efficient design platform is an organized data hierarchy that supports the coexistence and processing of algorithm analysis, simulation modules, and Building Information Models (BIM). The Building Information Modeling (BIM) database and environmental sensor network centrally collect all geometric, environmental, and operational data, and perform computational analysis with consistent parameterization. To support interoperability, this layer will comply with external standards of open BIM and data management systems.

The next step will involve repeatedly optimizing the building structure using a combined simulation-optimization algorithm. Based on dynamic energy and occupancy models, the simulation module will be used to calculate the heat transfer through the envelope, internal loads, and ventilation recovery. The above is the main part of energy forecasting.

$$E_{\text{hour}}(i) = U_e A_e \Delta T_i + Q_{\text{load}}(i) - \eta_v Q_{\text{vent}}(i) \quad (1)$$

where $U_e$ and $A_e$ are the overall heat transfer coefficient and area of the envelope, $\Delta T_i$ is the indoor-outdoor temperature differential for hour $i$, $Q_{\text{load}}(i)$ is internal gain, and $\eta_v Q_{\text{vent}}(i)$ is heat recovered from ventilation.

An easy-to-use interface is used to set the project's objectives, assumptions and regulations of the scenario, and to view simulation results. Figure 1 shows the system structure and typical workflow, demonstrating the interactions between BIM import, module data flow, multi-objective optimization, iterative simulation, and human-computer interaction analysis.

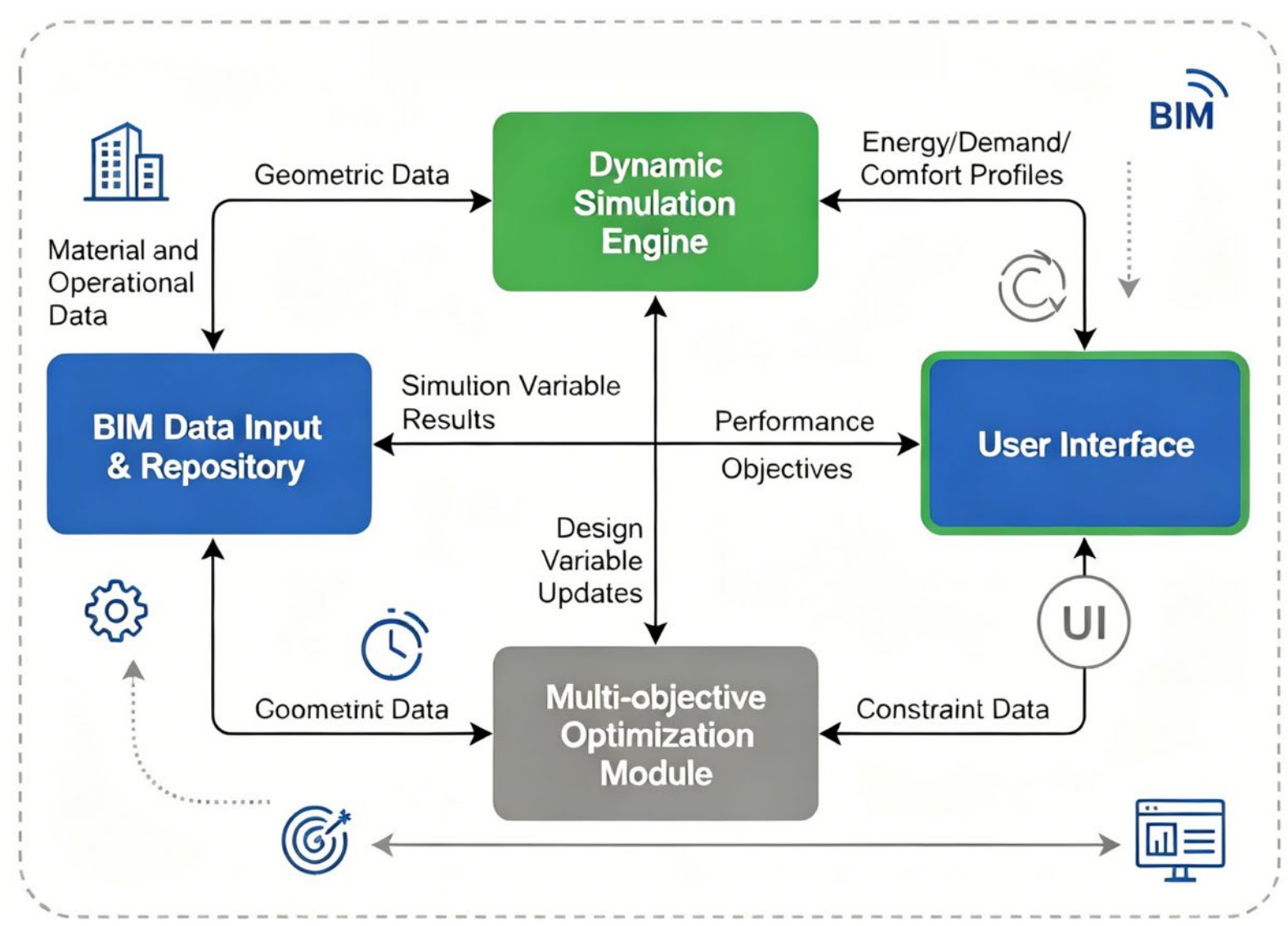

**Figure 1.** System architecture and integrated workflow of the proposed energy-efficient design platform.

*3.2. Algorithm Design and Platform Integration*

The core of the platform's analysis views energy-efficient building design as an optimization of multiple objectives under operational and regulatory constraints. The optimized solution consists of a vector of design parameters. These parameter vectors include the characteristics of the building envelope, HVAC systems, and control strategies, aiming to minimize the sum of annual energy consumption, lifecycle costs, and occupant discomfort.

The design issues can be listed as follows:

$$\min_{\mathbf{x}\in\mathcal{D}} [f_1(\mathbf{x}), f_2(\mathbf{x}), f_3(\mathbf{x})] \tag{2}$$

where $\mathcal{D}$ is the admissible design space, $f_1(\mathbf{x})$ denotes the estimated annual energy usage, $f_2(\mathbf{x})$ the discounted lifecycle cost, and $f_3(\mathbf{x})$ an aggregate discomfort score, often computed as a weighted sum of thermal and visual deviation metrics.

Dynamically calculate the annual energy consumption in a simulation by integrating the hourly predictions over the entire operating year:

$$f_1(\mathbf{x}) = \sum_{t=1}^{T} E_{\text{hour}}(\mathbf{x}, t) \tag{3}$$

where $E_{\text{hour}}(\mathbf{x}, t)$ is the predicted energy demand at timestep $t$, for given design vector $\mathbf{x}$.

A surrogate model was employed to train a kernel-based regressor using simulated samples of energy consumption along $\mathcal{D}$: to improve the speed of search.

$$\hat{f}_1(\mathbf{x}) = \sum_{i=1}^{m} \alpha_i K(\mathbf{x}, \mathbf{x}_i) \tag{4}$$

where $\alpha_i$ are model weights, $K(\cdot,\cdot)$ is the chosen kernel, and $\{\mathbf{x}_i\}$ are sampled designs.

The candidate solutions should meet all the practical and regulatory requirements listed in:

$$g_j(\mathbf{x}) \leq 0, j = 1, \dots, M \tag{5}$$

where each $g_j(\mathbf{x})$ is a constraint, such as a maximum energy demand, a minimum daylighting value or emission limit.

Using layer-based simulation-surrogate hybrid optimization, evolutionary algorithms are employed to evolve the candidate design population. The surrogate model quickly updates the population, and then fully simulates the promising solutions. The microservices architecture for infrastructure allows for the rapid development and deployment of high-performance, energy-efficient solutions that comply with current energy-saving regulations.

## 4. Implementation and Case Study

### *4.1. Platform Implementation Details*

The platform built has powerful computing performance and a multi-tier service-oriented architecture that supports scalability. This is a backend microservices architecture based on Python 3.11 and the Django REST framework, capable of executing multiple simulation and optimization tasks simultaneously. The high-performance modules of the computing core, such as the evolutionary optimizer and dynamic thermal simulation kernel, are written in C++ and connected to the Python layer via SWIG. PostgreSQL and PostGIS can easily store the large amounts of spatial and temporal data required for advanced architectural analysis, and they support data persistence.

React.js is used to create an interactive dashboard on the front end, which can display the progress of simulation and optimization phases in real-time. RabbitMQ message broker allows reliable event-driven communication between simulation, proxy modeling, and optimization services, and supports module synchronization and high-throughput data exchange. The platform will now be more open to developers, as an API gateway has been added to simultaneously support data access and security.

Functional modules include the BIM extraction interface for parsing geometry and material data; multi-source sensor data integrator; high-fidelity dynamic simulation engine; multi-objective evolutionary optimizer. Since the platform can display data and charts, it is possible to view the impact on electricity, expenses, and living standards at any time during the design process.

### *4.2. Case Study Description and Data Processing*

In order to test the platform's performance, a mid-rise office building (18,000 square meters, 12 floors) was selected in a temperate continental climate zone. The geometric data of the five different areas in the BIM model are assigned as areas 1 to 5. The main energy characteristics vary significantly in each area. For example, the ratio of wall area to window area in Zone 3 is 0.28, while in Zone 5, the ratio is 0.45. In Zone 5, the external wall U-value is the lowest at 1.0 W/m²K, while in Zone 3, the U-value is the highest at 1.5 W/m²K, indicating poorer thermal performance of the external walls. Every 100 square meters, the average occupancy density is 8.2 to 12.1 people.

All raw data have been standardized and analyzed. Using the Kalman filter for interpolation, the missing sensor records accounted for 3.2% of the annual time series; before manual validation, the multivariate outlier rate of the plug load profile averaged 2.6%. The cleaned data for all areas and relevant features were standardized into a single time series with a frequency of 15 minutes. Figure 2 shows the distribution of the three main driving factors (window-to-wall ratio, U-value, and occupancy rate) across the five areas, revealing different patterns. Area 2 has a high window-to-wall ratio of 0.41 and a relatively high occupancy rate of 11.0 people per 100 square meters, while Area 3 has a low window-to-wall ratio but high thermal conductivity of the envelope.

As shown in Figure 2, the aforementioned data-driven differences indicate that optimization and simulation must be calibrated at the regional level. This relatively detailed data can be used for scenario analysis and energy performance benchmarking studies.

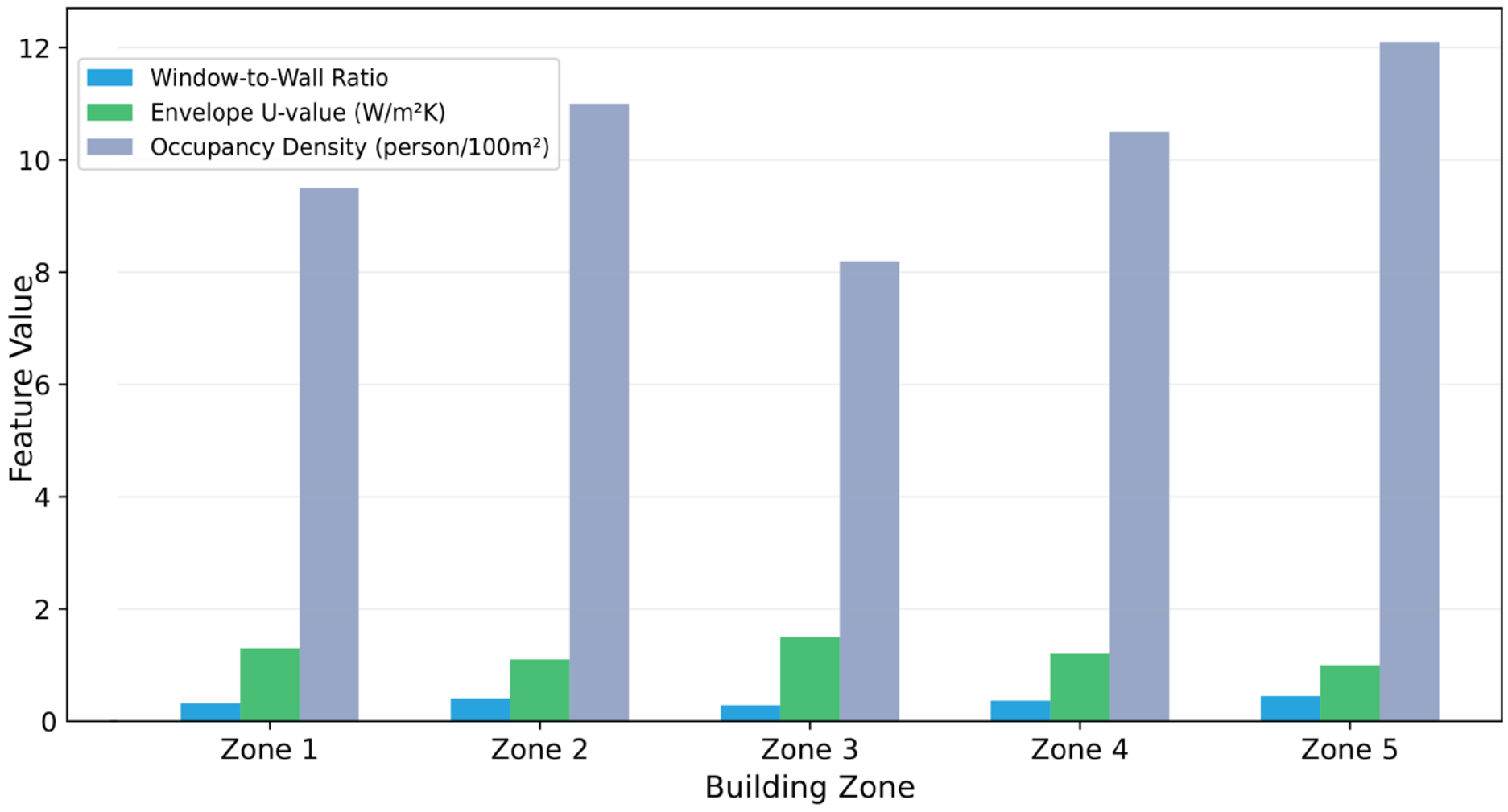

**Figure 2.** Feature distribution per building zone.

### *4.3. Experimental Results and Analysis*

40 years optimized the meticulously planned multi-regional dataset. As shown in Figure 3, in the initial stage, the total energy consumption of the building was 315 kWh/m·a. Between the 1st and the 15th iterations, this number significantly decreased. After the 30th iteration, the curve stabilized, reaching the limit of 223 kWh/m2 a. This is consistent with the results predicted by the platform's simulation and validation procedures, showing a decrease of 29.3% compared to the initial value.

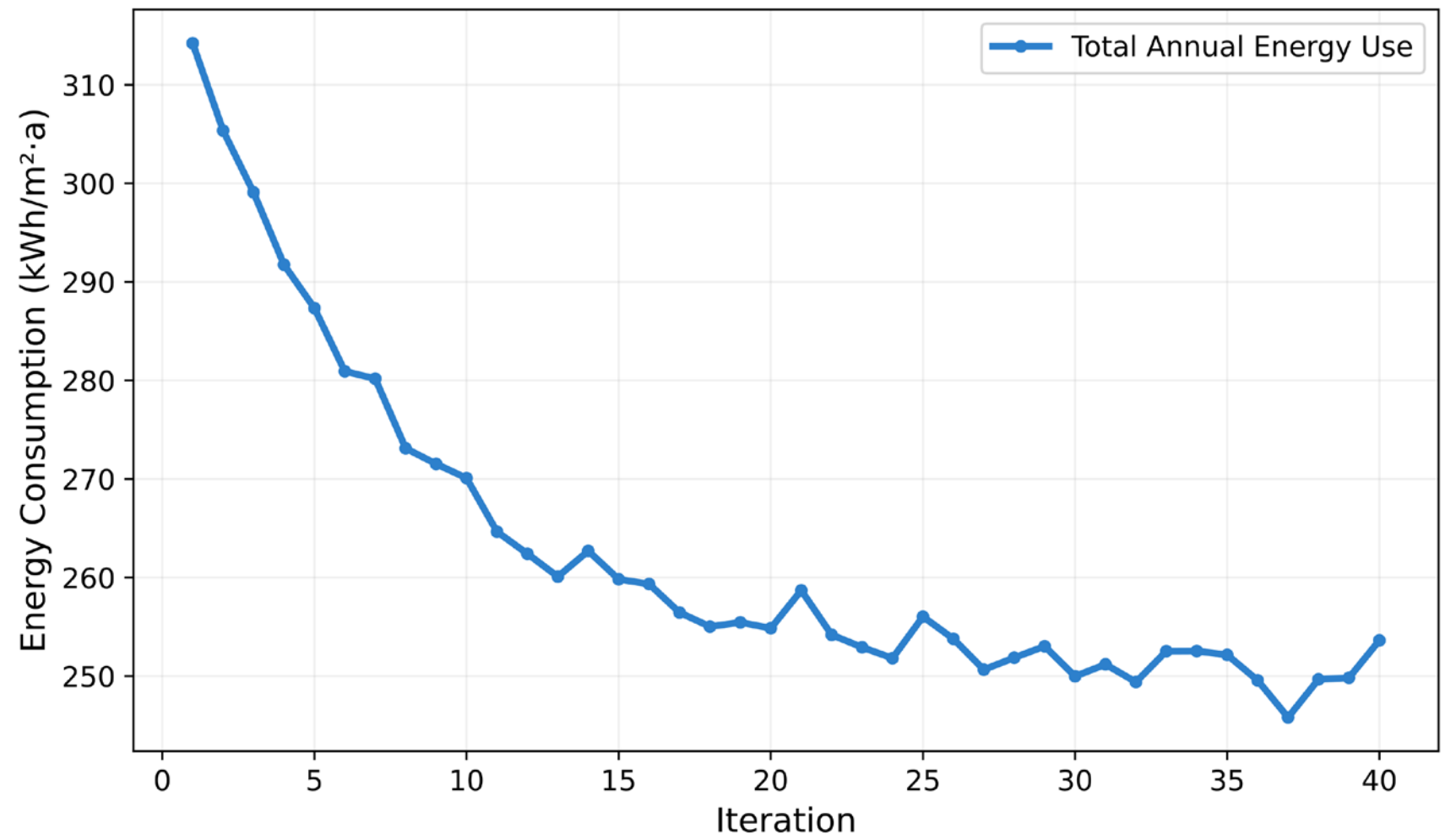

**Figure 3.** Optimization convergence curve.

Optimization improves people's lives by reducing energy consumption and costs. The weighted annual discomfort hours are below 70 hours, far below the legally mandated 110 hours. The lifecycle energy cost analysis indicates that the optimized solution set is economically feasible, as its lifecycle investment is only 3.7% lower than the baseline case.

Figure 4 shows the results. All Pareto-optimal design variables have been evaluated. The box plot shows that the optimal envelope U-value is between 1.05-1.57 W/m²K (with a median of approximately 1.23), and most nighttime ventilation rates are between 2.1-3.6 $h^{-1}$. According to the input features, the Pareto-optimal window-to-wall ratios are 0.28 and 0.45, with a median of 0.36. They are unevenly distributed spatially. The occupancy control index for adaptive scheduling of air conditioning and lighting is 0.37-0.89.

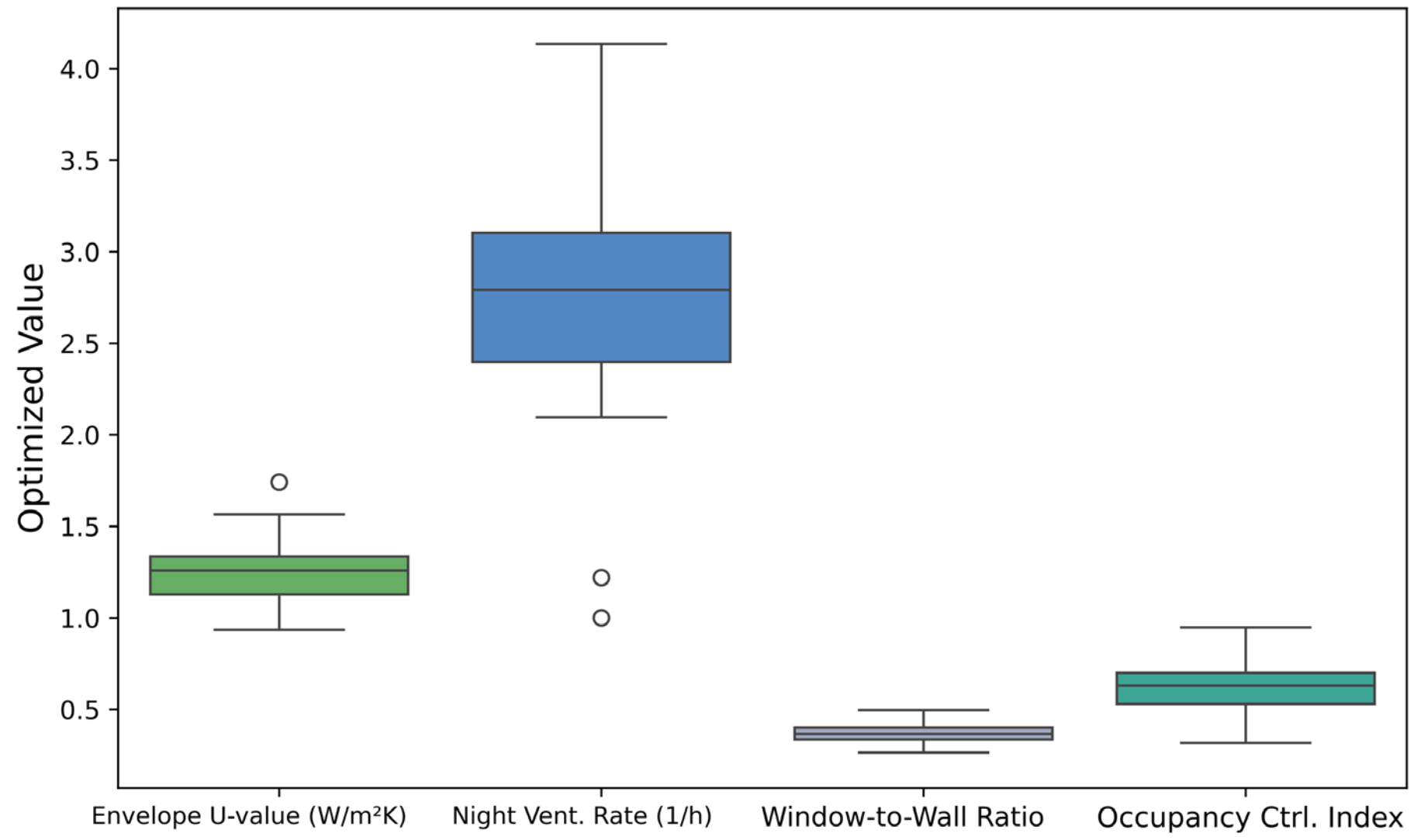

**Figure 4.** Boxplots for major parameters of Pareto designs.

According to the results of the correlation analysis, the nighttime ventilation rate and the window-to-wall ratio are positively correlated with the energy simulation results, with Pearson coefficients greater than 0.62. Based on the above results, the performance of the envelope structure and control parameters can be improved through appropriate adjustments without the need for significant financial investment. As shown in Figures 3 and 4, they together improve the convergence speed, while reducing result fluctuations during the platform's simulation optimization process, and maintaining a stable trade-off space.

## 5. Conclusion

This article introduces an engineering-centered green building energy design platform. The platform integrates advanced data extraction, high-fidelity simulation, multi-objective optimization, and a transparent service-oriented system. By integrating Building Information Modeling with the platform and actual operational data, comprehensive scenario design and decision-making are achieved. The three main extensions are: more closely integrating the C++ simulation core with adaptive agent models; stabilizing the handling of heterogeneous and missing sensor data; and a modular structure for dynamic feedback and iterative optimization based on real-time data visualization.

Using a typical multi-zone office building as a case study, the new system was found to be economically viable, as it can reduce annual energy consumption by 29% while maintaining relatively low costs. Full-spectrum parameter sensitivity analysis improves the accuracy and scalability of sustainable building design by introducing fine-grained data validation and parallel optimization workflows.

In real life, the proposed solutions will help architects, engineers, and decision-makers find high-performance, energy-efficient, and comfortable buildings, bridging the gap between the digital twin concept and building operations. In the future, the scope of uncertainty quantification will expand, more automation will be introduced in design input generation, and it will be applicable to more climate and operational conditions to support the large-scale construction of high-performance green buildings.